\documentclass[letterpaper, 10 pt, conference]{ieeeconf}  % Comment this line out if you need a4paper
\usepackage{graphicx}
\usepackage{float}

\usepackage{amsmath}
\usepackage{siunitx}
\usepackage{subcaption}
\IEEEoverridecommandlockouts                              % This command is only needed if 
\title{\LARGE \bf
Slip Detection and Surface Prediction Through Bio-Inspired Tactile Feedback
}

\author{Dexter R. Shepherd$^{1}$, Phil Husbands$^{2}$, Andy Philippides $^{2}$, Chris Johnson$^{2}$% <-this % stops a space
}

\begin{document}

\maketitle
\thispagestyle{empty}
\pagestyle{empty}

%%%%%%%%%%%%%%%%%%%%%%%%%%%%%%%%%%%%%%%%%%%%%%%%%%%%%%%%%%%%%%%%%%%%%%%%%%%%%%%%
\begin{abstract}

High resolution tactile sensing has great potential in autonomous mobile robotics, particularly for legged robots. One particular area where it has significant promise is the traversal of challenging, varied terrain. Depending on whether an environment is slippery, soft, hard or dry, a robot must adapt its method of locomotion accordingly. Currently many multi-legged robots, such as Boston Dynamic's Spot robot, have preset gaits for different surface types, but struggle over terrains where the surface type changes frequently. Being able to automatically detect changes within an environment would allow a robot to autonomously adjust its method of locomotion to better suit conditions, without requiring a human user to manually set the change in surface type.
In this paper we report on the first detailed investigation of the properties of a particular bio-inspired tactile sensor, the TacTip, to test its suitability for this kind of automatic detection of surface conditions. We explored different processing techniques and a regression model, using a custom made rig for data collection to determine how a robot could sense directional and general force on the sensor in a variety of conditions. This allowed us to successfully demonstrate how the sensor can be used to distinguish between soft, hard, dry and (wet) slippery surfaces. We further explored a neural model to classify specific surface textures. Pin movement (the movement of optical markers within the sensor) was key to sensing this information, and all models relied on some form of temporal information. Our final trained models could successfully determine the direction the sensor is heading in, the amount of force acting on it, and determine differences in the surface texture such as Lego vs smooth hard surface, or concrete vs smooth hard surface. Simpler approaches, such as average vectoring, can be used to track direction, but, unlike the neural model, cannot handle the complexities of surface classification.

%Environmental feedback is essential for a robot traversing over varied terrain. If an environment is slippery, soft, hard or dry a robot must adapt its method of locomotion accordingly. Current multi-legged robots such as the Boston Dynamics Spot robot have preset gaits for different surface types, but struggle over terrains where this surface type changes frequently. Being able to automatically detect changes within an environment would allow an agent to autonomously adjust its method of locomotion to better suit this, without requiring a human user to manually set the change in environment. Within this paper we make use of a bio-inspired tactile sensor known as a TacTip. We explored different processing techniques and a regression model, using a custom made rig for data collection to determine how an agent could sense directional and general force on the sensor. We further explore a neural model to classify specific textures. Pin movement (the movement of optical markers within the sensor) was key to sensing this information, and all models relied on some form of temporal information. Our end models could determine direction the sensor is heading in, amount of force acting on a sensor, and determine differences in the texture such as Lego vs smooth hard surface, or concrete vs smooth hard surface. Simpler approaches such as average vectoring can be used to track direction, but do not predict the complexities of classifying surfaces. This is where the neural model shows success. 

\end{abstract}

%%%%%%%%%%%%%%%%%%%%%%%%%%%%%%%%%%%%%%%%%%%%%%%%%%%%%%%%%%%%%%%%%%%%%%%%%%%%%%%%
\section{INTRODUCTION}

Insects and humans alike rely on some form of environmental feedback for adaptive behaviour, much of which is tactile~\cite{durr2014tactile, marieb2007human, krushynska2017spider,johansson1979tactile}. Tactile sensation supports adaptive behaviour for tasks such as locomotion and interaction. Such benefits would be useful for a robot. Walking over different surfaces - hard, soft, slippery, non-slippery - would produce different tactile sensations and require adaptation of the gait for efficient locomotion. For a robot to adapt its behaviour in real time each time the floor characteristics changed, it would need to be able to sense and classify these changes to enable changes in gait.  

Commercially successful walking robots such as Boston Dynamic's Spot have a series of settings that allow walking over a variety of surface types. However, the robot does not automatically detect the changes in the characteristics of the ground~\cite{wetzel2022use}. Indeed, the Boston Dynamics user guide provides a warning that changes within the environment surface types can cause the robot to fall over. This is because the robots rely mainly on visual input, something that many animals and insects do not need for traversal through challenging terrains because they use sophisticated tactile sensors. Detecting e.g. soft/slippery terrain using computer vision is very challenging. Other common techniques such as LIDAR do not provide information about the surface other than the shape~\cite{laser}. Many of the sensors that have explored slip detection and ground texture classification had shortcomings such as relying on moisture for slippage (not all surfaces are wet) or are limited to low velocity robots~\cite{giguere2011simple, patil2018design}. Much of the successful work in this area has been achieved by optical tactile sensors such as the TacTip. 

The development of the TacTip~\cite{lepora2021soft}, a skin-inspired optical sensor, has made detecting direction and pressure over large surface areas cheaper and easier. The TacTip does not suffer from mass wiring or physical gaps in sensing as each input is pixel distance apart. While there is a growing body of research on tactile sensors, there are still many challenges in developing agents that exploit tactile information for motor control. With more advanced tactile sensors such as TacTips, work has been done extensively to identify surfaces such as edges, and has gone further to identify objects from multiple touches. TacTips are skin-inspired soft tactile sensors that use cameras facing a soft dome-shaped body that can be used to detect touch forces~\cite{ward2018tactip}. The camera views a series of white dots on the inside of the mesh, known as optical markers, which move relative to each other according to the magnitude and direction of the touch force. The TacTips ”feel” by visual analysis of the camera image. Compared to arrays of pressure sensors, such vision-based tactile sensors have superior resolution, are relatively easy to fabricate, have a compact form factor, and require simpler multiplexing connections~\cite{10160877,yamaguchi2019recent}. TacTips have been used for improving grip on robots through edge and object detection~\cite{winstone2012tactip, lepora2021soft, church2021optical}. High-dimension imagery from these sensors allows a more accurate reading of touch and how much force/direction is applied. The morphologies of such sensors are not limited to specific shapes. Optical sensing in this way has been applied to larger surface areas for tasks such as balancing a leg~\cite{zhang2021tactile}, which successfully used a convolutional neural network to estimate and clarify ground slope. 

Though work on slip detection with tactile sensors exists~\cite{james2018slip} very little work has been done to detect slippage from surface type with the goal of walking over terrain. 

In this paper we carry out the first detailed investigation to assess the suitability of the TacTip for automatic detection of surface conditions as encountered in locomotion over challenging terrains. We explain the different techniques used for direction and force detection, and for surface classification, along with the details of  a custom made rig for data collection. We present a qualitative study of the sensor directional and pressure information, and how a neural model successfully predicted specific texture types.

\section{METHODS}
We developed two main approaches to classification tasks. One relies on pin marker tracking and changes within images, the other relies on a neural network. For data collection we made use of a custom-made rig that would run tasks over and over to give us plenty of directional and impact data to make models with. In addition to volume  of data, the repeatability of such conditions is achievable using this rig, compared to human data-gathering. 

\subsection{TacTip Construction and Processing}
Our TacTip featured a 3D-printed chassis housing a USB webcam with a wide-angle lens. Unlike standard approaches using gel for touch sensitivity, our sensor used clear silicone, which is less sensitive but more durable. The LED ring illuminates the optical markers but caused reflections on the silicone, complicating pixel intensity thresholding. To address this, we removed large binary blobs beyond a certain size threshold, with 1cm equating to roughly 100 pixels.

We focused on two key aspects: pressure/touch (using a button sensor) and directional forces. Achieving directional force sensing proved challenging. We explored mathematical and regression-based approaches to determine the best method for sensing directional forces.

A standard approach of preprocessing optical sensors is to binary threshold the image (figure~\ref{fig:processingTip}).  This made use of an adaptive threshold. 

\begin{figure}
\centering
  \includegraphics[clip,width=1\columnwidth]{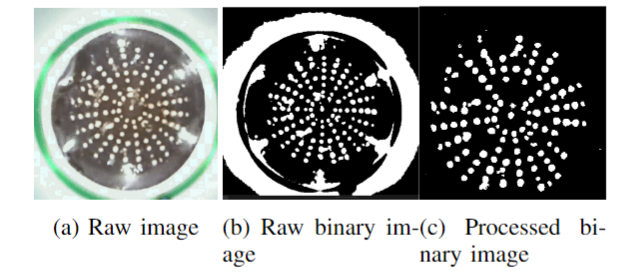}%

\caption{The raw binary image is then processed through the removal of the larger blobs. Still, the glare has some effect but reduces the complications of too much white, such as clustering in the middle. }
\label{fig:processingTip}
\end{figure}

\subsubsection{Contact detection}

Force results allow an agent to understand what force it is getting from an environment, and where on the sensor it is getting this force. For the task of classifying different surface types, we needed to create a large data set of tactile information where contact with the surface was recorded and non-contact was not. Non-contact would look the same whether we are recording slippery surfaces or hard surfaces, therefore must be removed from the data. 

Our method of calculating contact from the binary image made use of a receptive-field-inspired mechanism applied as a grid over the tactile image. We took the absolute difference between two TacTip image frames (one after the other) and summed all the pixels in each grid square. If our grid size is 5 then the tactile image will be viewed as 5 $\times$ 5 images. 
 
The force within each grid point (denoted in a matrix that starts empty and recursively enters this function $P$) calculates the difference between the frames $F$ at each index within the selected grid square. The bounds are set by dimensions x and y ($d_x,d_y$). $I$ represents the initial index to show the original grid. We subtract the global average from the change in the frame to highlight a more significant change, thus removing noise. Finally, we subtract $\gamma$ which represents a temporal dampener that reduces pixels over time. This is how we get the faded effect as a stimulus drops (figure~\ref{fig:touchTip}).  

 \begin{equation}
     P_{j:d_x,i:d_y} = \frac{\sum_{j}^{d_x}\sum_{i}^{d_y} \left | F_{t,i,j} - F_{I,i,j}\right | }{(d_x-j)*(d_y-i)} - \gamma -\frac{\left | F_{t} - F_{I}\right |}{n}
    \label{force}
 \end{equation}

 \begin{figure}
\begin{center}
  \includegraphics[clip,width=0.8\columnwidth]{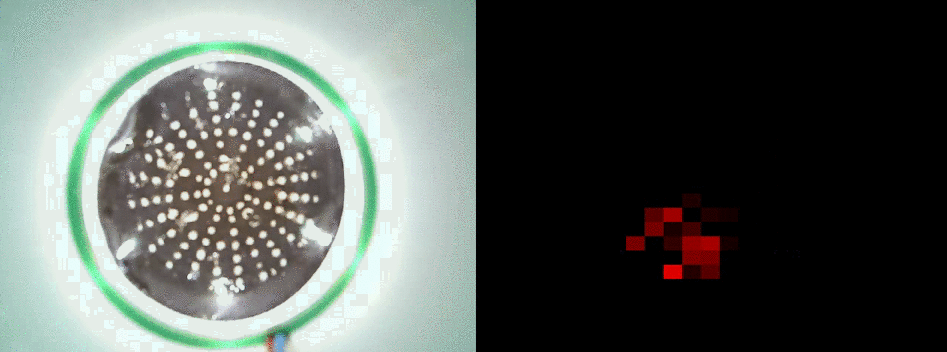}
  \caption{The stimulus is presented in the bottom left corner. This movement has left a fade around the area, showing this is not the contact point. It has already had initial contact, and now the touch is moving away from the TacTip. }
  \label{fig:touchTip2}
\end{center}
\end{figure}

\subsubsection{Direction}

We compared two point tracking methods for directional detection: one employed a regression model trained on human-labelled data, while the other utilised a vector mapping function with binary thresholds for point differentiation. The former required no additional preprocessing, making it efficient for reducing input data size into a large surface classification model.

Post-preprocessing, we collected centroid points~\cite{zhang2018fingervision} for each blob to monitor general movement. Our model estimates point movement between frames, beginning from an origin frame during object initialization, aiming to maximize the points detected. This function operates on binary images converted into arrays of centroid points (n,2).

Next, we apply a recursive mapping function to find the shortest distance between points, calculated using the Euclidean distance formula between origin points (o) and mapped points (t). Matrices are passed through the equation to calculate all distances.

\begin{equation}
    d(o,t) = \sqrt{\sum_{i=1}^{n}(o_i-t_i)^2}
    \label{direction}
\end{equation}

Due to LED ring noise, tracking TacTip points can yield variable point counts. To address this, we manually labelled a TacTip point data set, augmenting it through translation, zooming, and then employed a Ridge regression model (alpha=150) for classification. This model effectively tracked TacTip movement, resilient to noise, while the vector method excelled in tracking point disturbances but was noisier.

\begin{figure}
\centering

  \includegraphics[clip,width=1\columnwidth]{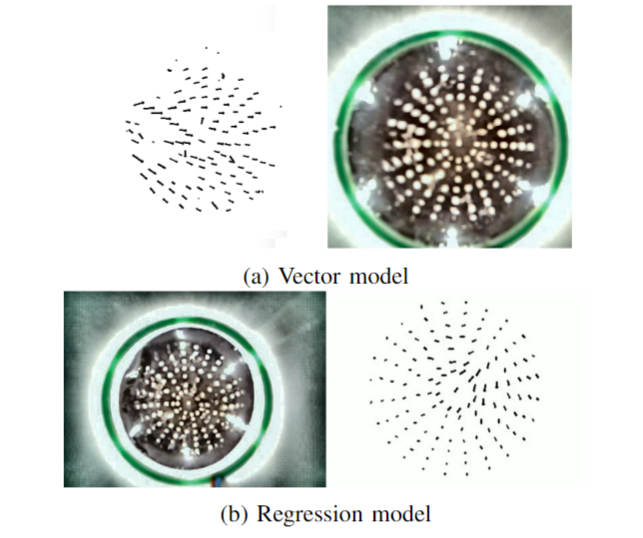}%

\caption{The two models predicting/localising the optical markers. The regression model will always have 133 vectors. The vector model will vary. \emph{a} shows a drag in a direction towards the right of the image. The vectors show an outward direction, with some noise. For \emph{b} there is a pressure placed on the top right of the TacTip causing an inward force within the vectors. }
\label{fig:processingTip}
\end{figure}

\subsection{Rig}
After preprocessing there is much information the TacTip sensor can provide. With the preprocessing technique that plots the movement of vectors, discussed in a previous section, we can get the average vector. This represents the overall direction being acted on the sensor. We can additionally gather pressure from separate receptive fields, or sum all the receptive fields to get an overall analogue sense. These aspects of the preprocessing techniques are explored within this section using different actions and surfaces. The section evaluates how accurate the methods we have developed are at extracting useful information.

\begin{figure}
\begin{center}
   \includegraphics[clip,width=1\columnwidth]{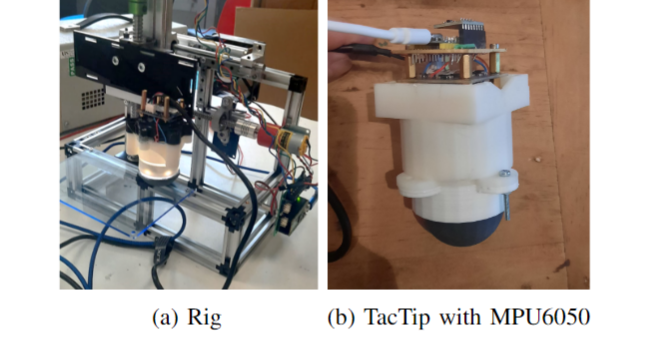}

  \caption{\emph{a}: The rig used to have a 2-axis system to gather data from the sensor. This method used a stepper motor and DC motor and ran using a Raspberry Pi Pico using serial communication with a PC that would send movement instructions and gather data. \emph{b}: An  MPU6050 (accelerometer and gyroscope sensor) is attached to a RP2040 microcontroller running MicroPython. This variant connects via serial to a machine that is also connected via USB to the camera of the TacTip. It simultaneously records from both. In the future having both camera and gyroscope on the same cable and PCB would be ideal. }
  \label{fig:rig}
\end{center}
\end{figure}

We built a repeatable testing rig (the rig) to eliminate noise and human error, allowing for continuous, unattended data collection. The rig, as shown in Figure~\ref{fig:rig}, featured a base plate with a pressure sensor beneath the TacTip for assessing pressure variations. We swapped the base for different experiments, including a foam plate to test softness and a water-containing tray for slippage tests.

Constructed from aluminum profiles and equipped with stepper motors, the rig facilitated various trials, each running hundreds of times for data generation, as detailed in the results section.

\subsection{Neural Model}
We developed a classifier that would take in the vectors from $T$ frames and flatten them into a linear input layer of a network. The network had an output layer of four where each neuron determined a surface property. The network used a neuron dropout rate of 0.2 and made use of Sigmoid activation functions on each layer (excluding the final layer). $T$ provides some temporal context for the model. 

Our initial data set was made from a prerecorded video going over different floor states. In order to filter out insignificant information (as in not touching anything) we used the force detection method described in equation~\ref{force}. All frames where touch was present had $T$ amount of frames prior to the current frame concatenated into one image. We initially took the points from the frame, but inaccuracies within the marker prediction prevented classification of some textures. When concatenated together the model had an input of $(n,h*T,w)$. The labels were labelled in the format \{soft hard slippery no-touch\}. The data was scaled down using a standard scalar that removed the mean $\mu$ and scaled to a unit variance $\sigma$. The standard score of a sample $x$ is calculated as:
\begin{equation}
    z = (x - \mu) / \sigma
\end{equation}

The neural architecture was made up of two hidden layers, five outputs and an input layer determined by the time step size.  Using the vectors as input was not always accurate for smaller movements caused by the texture of a surface. This led to the development of an architecture that took in binary threshold images concatenated together for $T$ frames. 

\begin{figure}
\begin{center}

  \includegraphics[clip,width=1\columnwidth]{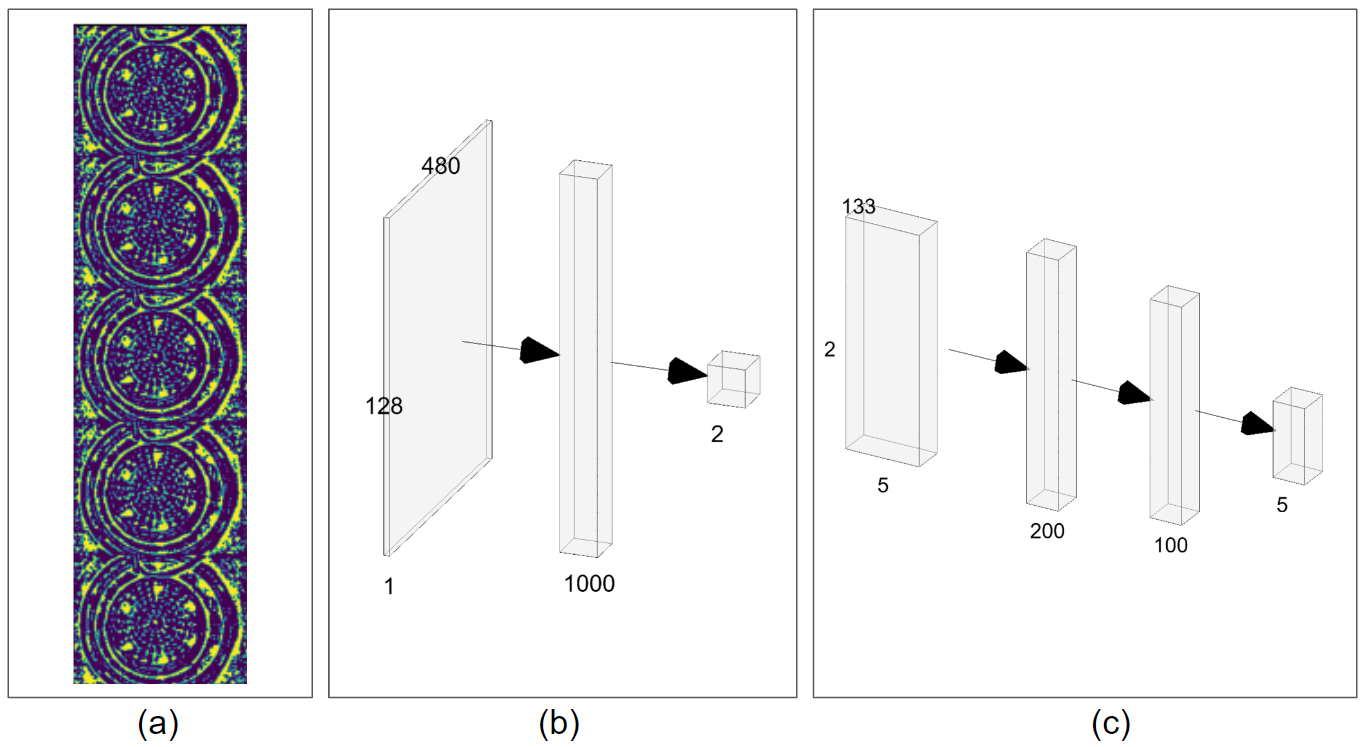}

  \caption{A diagram showing an example neural architecture. This is an example of the architectures with arbitrarily chosen layer sizes. Alternatively, the classifier could have multiple classes such as soft, hard, and slippery to classify which one the surface is, which is what we chose as our classifier. \emph{b} is a 2D convolutional neural network using a filter size of $8 \times 8$ and a single hidden layer of 1000 nodes. The input is represented in \emph{a} as binary images over $T$ frames concatenated together. \emph{c} is the FNN structure designed for a series of vectors. }
  \label{fig:surfaceModelArch}
\end{center}
\end{figure}

\section{RESULTS}

\subsection{Direction}

Different forces, directions and speeds can be detected through the TacTip sensor. Action detection is important for a robot so it can establish whether a leg is placed down, a surface is moving (and the leg is not) and what the shape of the force is. 
\begin{figure}
\begin{center}

  \includegraphics[clip,width=1\columnwidth]{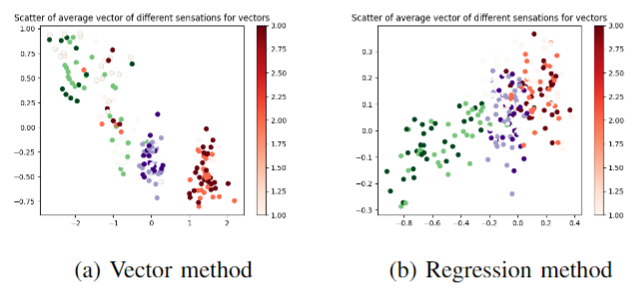}

\caption{Direction of different sensations trialled 30 times on each sensation, plotted as points in 2-dimensional space. This was trialled on both direction detection models. The shades represent the pressure that a surface physically interacted with (as an approximation) then split between 0 and 1, where 0 was very little pressure and 1 was a full press onto the surface. The exact measurements are not required as this is to demonstrate how more weight affects the points. There is a linearly separable relationship between the left and right acting forces. \emph{Green} shows left acting force, \emph{red} shows right acting force and \emph{blue} shows a central push. On the regression model, we can see that the distances from each trial are smaller (looking at the axis) than on the vector model. }
\label{fig:all_dirs}
\end{center}
\end{figure}

With minor errors, such sensations are grouped together and can easily be classified using the average vector force with simple models. A push on the centre creates a small magnitude after averaging the vectors. This is visible within figure~\ref{fig:all_dirs}. 
The latter shows the directional vector x-axis is around 0 which is expected if the points are being pressed in the centre of the sensor. There is a larger spread of values within the y-axis. Inspection of which dots are found and which are not shows that fewer dots can be visible on one side once again due to glare. The higher force markers are typically more on the outside with the regression model, compared to the vector model. This is likely due to more noise interference with vectors using the vector model. 

%TODO ind STD of the spread of values. 

% \begin{figure}
% \begin{center}
% \subfloat[push]{%
%   \includegraphics[clip,width=0.5\columnwidth]{Assets/Results/save24.png}
% }
% \subfloat[forward]{%
%   \includegraphics[clip,width=0.5\columnwidth]{Assets/Results/save16.png}
% }
%   \caption{\emph{a}: A qualitative inspection of the force from a flat surface on the centre of the TacTip. \emph{A} shows what the camera is seeing, \emph{B} is the preprocessing step to gather the vectors. We then plot the averaged vector as a direction in \emph{D}. \emph{b}: Another example of the TacTip, but the sensor is bring dragged along a surface. The average vector successfully predicts the movement of the direction.}
%   \label{fig:qualskinDir}
% \end{center}
% \end{figure}

Though figure~\ref{fig:all_dirs}b has some overlap between left and right, it can be seen that the more force that acts, the less confused it is between. Pushing the sensor directly onto a flat surface (labelled as the top in figure~\ref{fig:all_dirs}a) is less clear within this classification. Upon qualitative inspection, we can see that the push of the sensor gives vectors outwards from the sensor. The overall output pushes downward, as the average is made up of all the velocities. If the sensor is at a slight angle this can have an effect on the vector. The magnitude of such a vector is much lower than when a force is being applied on one of the sides. We see this in most the 'Top' labelled examples within figure~\ref{fig:all_dirs}a. 

The method of gathering pressure described in equation~\ref{force} is good at locating touch, however, does not offer accurate depth of pressure. Using the vector direction tracking we can get the sum of all magnitudes from the resultant vectors $V$ for $n$ number of points. The equation below denotes how each value is gathered:

\begin{equation}
    P =\sum^{n}( \sqrt{\sum^{2}(V_x^2 + V_y^2)}  )
\label{overall_mag}
\end{equation}

\begin{figure}
\begin{center}
\begin{tabular}{||c c c c c c||} 

 \hline
 & 0.0cm & 0.2cm & 0.4cm & 0.8cm & 1cm \\ [0.5ex] 
 \hline\hline
 Average & 52.01 & 50.48 & 50.59 & 51.08 & 51.3 \\
 \hline 
 Median & 52.79 & 45.52 & 57.84 & 56.99 & 54.76 \\
 \hline
 Range & 42.82 & 36.9 & 32.33 & 32.58 & 31.39 \\ 
 \hline
 \end{tabular}
\caption{This table presents the sum of magnitudes $P$ from a pressure map as described by equation~\ref{direction} and processed equation~\ref{overall_mag}. These values were acquired over 100 trials of pushing the sensor multiple depths on a hard surface. 
 }
\label{table:dataAVGSTD}

\end{center}
\end{figure}

There is a positive trend with increasing pressure and total magnitudes. The sensor is being pushed further, therefore the optical markers are moving further outwards. However, it was found that there was a significant range between the magnitudes and there was not a positive trend in every trial. Using overall magnitude is not accurate enough to predict specific pressure. The task itself is not as linear and straightforward as a line of best fit. 
The model used would take in all vectors and thus provide more directional information across the TacTip. If all vectors are considered we can accurately predict pressure using linear regression.

If all vectors are considered we can accurately predict pressure using a regression model. On unseen data, the predicted values had an average difference of 1.25. 

\begin{figure}
\begin{center}

  \includegraphics[clip,width=1\columnwidth]{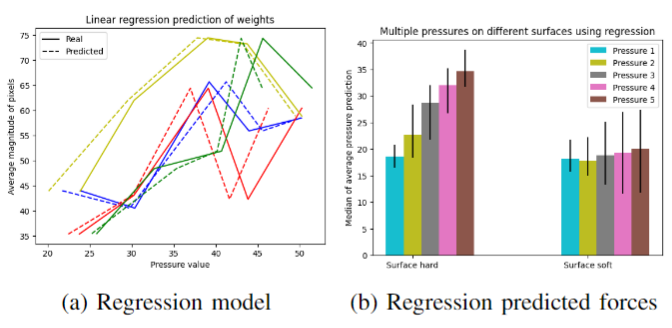}

\caption{\emph{a}: The predicted vs real pressure values from the ridge regression model. This represents a larger amount of noise part of the data set, where a larger number of optical markers are used. The model is able to predict pressures based on the relationship between individual marker magnitudes and the force itself. \emph{b}:  A plastic hard surface and a foam soft surface (supported by a hard surface base) were trialled against each other. The sensor is lowered to the point of touch and different pressure settings are evaluated. As the setting increases, the sensor is lowered further onto the surface. The predicted force is plotted, and we display the median of 100 trials on each pressure setting. The range from max to min value is represented for each class as a black line. }
  \label{fig:linReg}
\end{center}
\end{figure}

The pressure value is taken from a pressure sensor added to the plate that the TacTip is pushing against. Using analogue readings of the sensor we could gather how much force is being applied on the surface. 

%A trend within every trial is an increase in number of pixels. This is due to the fact the optical markers are moving from the original frame. We experimented with receptive field grids of $5 \times 5, 10 \times 10, 15 \times 15$ and $20 \times 20$. The detail increases with n. The $5 \times 5$ has much larger receptive fields than the $20 \times 20$, therefore the $20 \times 20$ is more sensitive to touch. 

The regression model was key to pressure detection as our first attempt tried to simplify the process by getting the average magnitude of vectors. The further the points are out, the more pressure that is being applied. This was true for light pressures, however did not work for distinguishing larger forces acting on the sensor.

\subsection{Simple Surfaces}
The state and morphology of a surface will read differently on the TacTip when the same pressures/ directions are applied to the sensor. It is useful for an agent to understand the surface it is interacting with so that it can adapt the way it interacts. 

\begin{figure}
\begin{center}

  \includegraphics[clip,width=1\columnwidth]{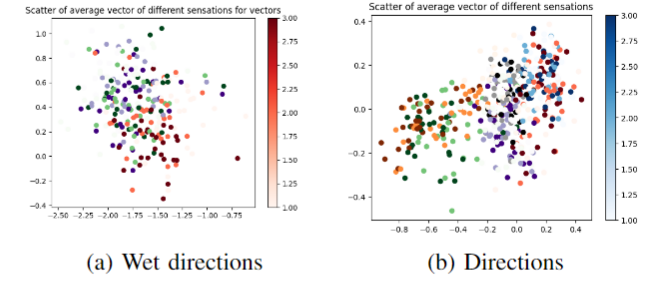}

  \caption{\emph{a}: We used a wet surface and the sensor was moved in the left, right and centre forces over both surfaces. There is less resistance from the slippery surface making the directional force have less of a significant impact if we compare these results to figure~\ref{fig:all_dirs} where the movement is distinguishable.
  \emph{b} shows the same experiment as in figure~\ref{fig:all_dirs} but both the soft and hard surfaces the sensor is touching. Each time the sensor is lowered to the point of touching the surface. The experiment on the soft body is shown in shades of purple (front press), green (left) and red (right). The hard body is shown in grey (front press), orange (left) and blue (right). It is clear that there is little way to use the overall magnitudes to distinguish what surface we are pressing. There seems to be minimal difference.
  }
  \label{fig:differentObj}
\end{center}
\end{figure}

The resistance of the material makes a significant difference in both force and direction. Figure~\ref{fig:linReg}b shows that the softer materials that offer less resistance are detected by the TacTip and thus pressure prediction on the soft body has a much lower value than the hard counterpart. The sensor response is near indistinguishable for soft and hard sensations when the pressures are similar. As pressure increases there is a significant gap between the pressure predictions which used the same method shown in figure~\ref{fig:linReg}a. 

Classification of a surface can not rely on pressure reading alone. Knowledge of how much pressure the system thinks it is putting down is required for classification of soft/hard surfaces. Figure~\ref{fig:linReg} highlights this where higher force on a soft material is similar to smaller pressure on a hard material. Wet surfaces can make a difference as evident in figure~\ref{fig:differentObj}a. If a robot is expecting movement and receives none, then it could mean the surface is wet. There is a clear difference between wet trial data and dry trial data, showing that the sensor cannot accurately predict direction on a slippery surface. 

%%%%%%%%%%%%%%%%%%%%%%%%%%%%%%%%

\subsection{Complex Surface Classification}
Though we can detect simple properties such as soft vs hard, slippery vs non slippery using the pin tracking methods, these models cannot distinguish between similar textures such as Lego and other hard surfaces. We employ the use of a CNN to predict the differences between textures.

\begin{figure}
\begin{center}
  \includegraphics[clip,width=1\columnwidth]{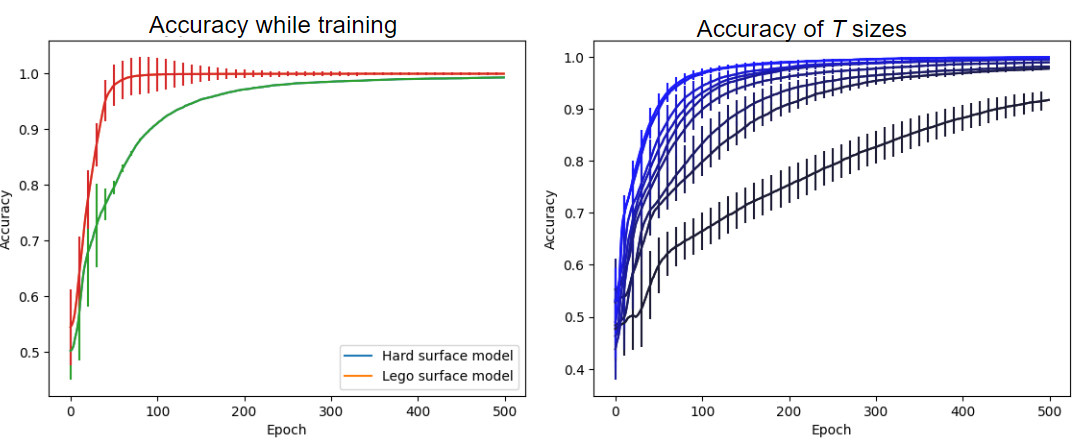}

  \caption{The training process of the model using stochastic gradient descent. When trialled on unseen data it predicted 99\% correct on unseen data. We then trialled it on one to ten values of T and found that the larger the value of T, the greater the accuracy. Within the figure labelled "Accuracy of T sizes" the darker colours represent lower values of T, whereas the lighter values represent higher values of T. Each T was averaged over 20 trials. The task was a classification task between soft, hard, slippery and no touch at all. These both used a CNN architecture with a dropout rate of 0.2. The error bars in the left plot represent the standard deviation at each epoch. The averages plotted used 20 trials on each model. }
  \label{fig:sgdAndTsize}
\end{center}
\end{figure}

The image-based network was trained on a texture data set to distinguish the difference between smooth and rough. The rough data was the sensor moving over a Lego base. When inspecting the video that makes up the data set we noticed the only real difference is some of the white dots shake slightly more on Lego. This classification task was performed at 99.9\% accuracy and averaged over 20 trials. These small movements rely on previous knowledge of movement in order to classify how much these frames have moved. When venturing to classify the difference between concrete and smooth wood the sensor performed at 99.3\% accuracy averaged over 20 trials. Both models used a convolutional neural network with the input as the entire image rather than the vector features. The vectors reduced the dimensionality, thus decreased memory demand of the model, however issues in the accuracy of predicting the markers would be amplified by real-time testing of the sensor. We used an adaptive binary threshold of the input images to make the optical markers stand out. The results of this experiment are seen in figure~\ref{fig:sgdAndTsize}. 

A higher value of $T$ showed improved performance on the model, likely because it provided better temporal context. The highest accuracy of these models achieved 100\% accuracy on the train data set where $T$ was set to 10 frames.

\section{CONCLUSIONS}
Our TacTip was able to detect directional force, contact force and classify specific textures. Directional force could be detected using optical marker tracking from original points to current points. Such points provided a series of vectors that could provide an overall averaged vector displaying which direction the sensor is moving in. We developed further models that would take in all vectors and predict the amount of pressure placed upon the sensor. When detecting pin markers data using the regression point predicting model was less affected by noise thus leading to more accuracy as an overall reading. With the evaluation of the experiments from this paper, we developed a model that classifies surface information. Using the same sensor for pressure and surface state is an unexplored avenue, and additionally makes efficiencies in design by having a singular sensor. The interpretation of this data was benefited by the construction of the test rig that allowed us to gather large quantities of data without human interference. 

For complex surface prediction such as Lego vs hard smooth surfaces the difference in the raw video is tiny movements within the optical markers. The use of a neural model to classify this information was effective for surface types and specific classification of similar surfaces such as Lego vs. flat hard surfaces. Prior information was additionally not needed, unlike it was for the vector-based classification tasks performed by the rig. 

Though our neural models are predicting between two surfaces, preliminary work has shown that we can predict between multiple surfaces. 

This work has ventured into small and simple to construct sensors that can detect different surfaces and textures. Such work will support the building of multi-legged robots that can adapt their gait accordingly to surfaces. This is something that many multi-legged robots struggle to do and is something we hope to achieve in future work.

\addtolength{\textheight}{-12cm}   % This command serves to balance the column lengths

%%%%%%%%%%%%%%%%%%%%%%%%%%%%%%%%%%%%%%%%%%%%%%%%%%%%%%%%%%%%%%%%%%%%%%%%%%%%%%%%

%%%%%%%%%%%%%%%%%%%%%%%%%%%%%%%%%%%%%%%%%%%%%%%%%%%%%%%%%%%%%%%%%%%%%%%%%%%%%%%%



\section*{ACKNOWLEDGEMENT}
This work was funded by the Leverhulme trust and EPSRC

%%%%%%%%%%%%%%%%%%%%%%%%%%%%%%%%%%%%%%%%%%%%%%%%%%%%%%%%%%%%%%%%%%%%%%%%%%%%%%%%

\end{document}